\pdfoutput=1

\documentclass[11pt]{article}

\usepackage[]{EMNLP2022}

\usepackage{times}
\usepackage{latexsym}

\usepackage[T1]{fontenc}

\usepackage[utf8]{inputenc}

\usepackage{microtype}

\usepackage{inconsolata}

\usepackage[para]{threeparttable}
\usepackage{booktabs}
\usepackage{multirow}
\usepackage{graphicx}
\usepackage{amsmath}
\usepackage{amssymb}
\usepackage{hyperref}
\usepackage{caption}
\usepackage{subcaption}

\newcommand\blfootnote[1]{%
  \begingroup
  \renewcommand\thefootnote{}\footnote{#1}%
  \addtocounter{footnote}{-1}%
  \endgroup
}

\title{Saving Dense Retriever from Shortcut Dependency \\ in Conversational Search}

\author{
    Sungdong Kim$^{1,2}$ \quad Gangwoo Kim$^{3\diamond}$ \\
    NAVER AI Lab$^{1}$  KAIST AI$^{2}$  Korea University$^{3}$ \\
    \texttt{sungdong.kim@navercorp.com} \quad \texttt{gangwoo\_kim@korea.ac.kr}
}

\begin{document}
\maketitle
\blfootnote{\textsuperscript{$\diamond$} Work done while interning at NAVER AI Lab}
\begin{abstract}
Conversational search (CS) needs a holistic understanding of conversational inputs to retrieve relevant passages. In this paper, we demonstrate the existence of a \textit{retrieval shortcut} in CS, which causes models to retrieve passages solely relying on partial history while disregarding the latest question. With in-depth analysis, we first show that naively trained dense retrievers heavily exploit the shortcut and hence perform poorly when asked to answer history-independent questions.
To build more robust models against shortcut dependency, we explore various hard negative mining strategies. Experimental results show that training with the model-based hard negatives~\cite{xiong2020approximate} effectively mitigates the dependency on the shortcut, significantly improving dense retrievers on recent CS benchmarks. In particular, our retriever outperforms the previous state-of-the-art model by 11.0 in Recall@10 on QReCC~\cite{anantha2021open}.\footnote{The code is available at \href{https://github.com/naver-ai/cs-shortcut}{github.com/naver-ai/cs-shortcut}.}

\end{abstract}

\section{Introduction}

Conversational search (CS) is a task of retrieving relevant passages from a large amount of web text given the current question and its conversational history, which consists of previously asked questions and their answers~\cite{dalton2019cast}. 
Unlike open-domain question answering (ODQA) taking a single question~\cite{voorhees-tice-2000-trec, chen2017reading}, CS assumes a sequence of questions interactively taken from information seekers. Hence, the questions need to be understood with the conversational history to find relevant evidence at each turn.

\begin{figure}[t] 
\centering
\includegraphics[width=0.5\textwidth]{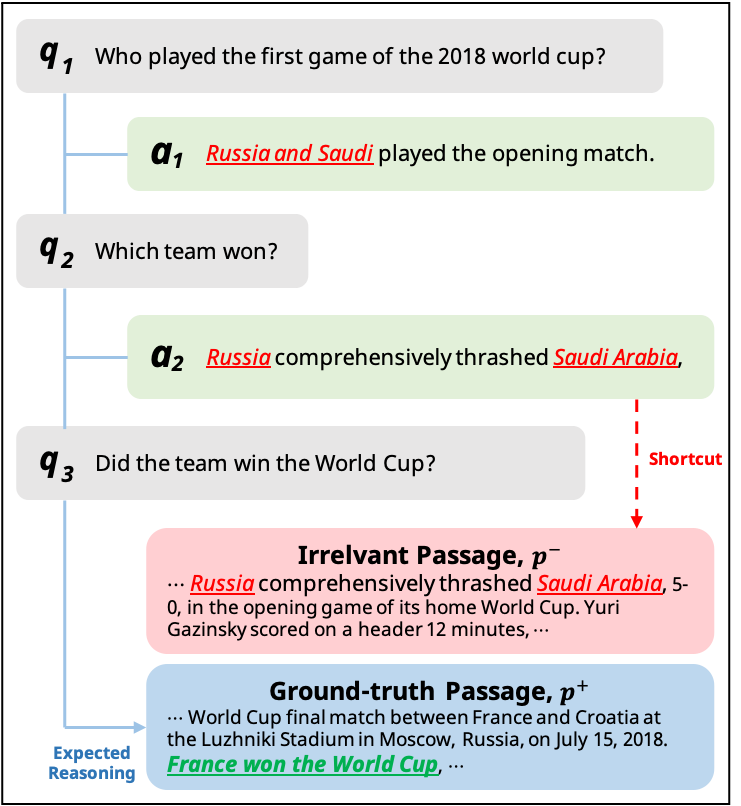}
\caption{An example of a retrieval shortcut in conversational search. While we expect the retriever to predict relevant passages by using all conversational inputs up to $q_3$ (Blue solid line), a dense retriever often ignores current turn question $q_3$ and only exploits previous history, $a_2$ (Red dashed line). We show the shortcut dependency is harmful to robust retrieval.}
\label{fig:shortcut_example}
\vspace{-5mm}
\end{figure}

To build a retriever that properly makes use of the conversational history, we first analyze a simple dense retriever baseline trained on one of the CS datasets, QReCC~\cite{anantha2021open}.
Our analysis shows us the existence of a retrieval shortcut in recent CS datasets, indicating dense retrievers heavily rely on the shortcut and retrieve irrelevant passages. Specifically, these shortcuts represent the spurious correlation between the conversational history and the relevant passages, pushing the dense retrievers to ignore current questions. For example, as illustrated in Figure~\ref{fig:shortcut_example}, a dense retriever retrieves wrong passages only paying attention to `\textit{Russia}' and `\textit{World Cup}' mentioned in the previous history ($a_1, a_2$) while ignoring the crucial cue `\textit{win the World Cup}' in the current question $q_3$.

Motivated by our observation, we further test how much the shortcut contributes to the performance of current retrievers. First, we build a simple BM25 baseline, which only takes the previous conversational history as input, but still performs surprisingly well on QReCC. Similarly, a dense retriever trained by feeding the conversational history without the current question keeps 70-80\% of the original performance. It implies a significant effect of the shortcut dependency on dense retrievers. From our analysis, we find the shortcut is more likely to be learned when the topic of conversation is constant. In other words, performance of the models drops especially when they are asked to answer history-independent questions.

To alleviate the overreliance on the shortcut, we explore using hard negative mining strategies, which have been recently proposed in ODQA and CS~\cite{karpukhin2020dense, xiong2020approximate, yu2020fewshot, lin2021contextualized}.
Experimental results show the model-based hard negatives make remarkable improvements in various CS benchmarks and are especially helpful to history-independent questions, mitigating the dependency on the shortcut effectively. Our retrievers outperform baselines by 11.0 in Recall@10 on QReCC.

Our contributions are summarized in three folds:
\begin{itemize}
    \setlength\itemsep{0em}
    \item We reveal the presence of a \textit{retrieval shortcut} in the conversational search, and dense retriever dependent on the shortcut is poor at generalizing toward a real scenario.
    \item We show training the dense retriever with hard negatives effectively mitigates the heavy shortcut dependency by in-depth analysis.
    \item We achieve a new state-of-the-art of recent CS benchmarks, QReCC and OR-QuAC.
\end{itemize}

\section{Background and Related Work}

Let $X_t = \{q_1, a_1, ..., a_{t-1}, q_t\}$ is a conversation up to turn $t$ where the $q_t$ and $a_t$ are the question and answer at turn $t$. We assume pre-chunked passages collection $\mathcal{C} = \{p_1,p_2,...,p_{|\mathcal{C}|}\}$ for the retrieval. Then, the formal objective of conversational search is learning function $f: (X_t, \mathcal{C}) \rightarrow P_t$, where the $P_t = \{p_1, p_2, ..., p_k\} \subset \mathcal{C}$ and $k \ll |\mathcal{C}|$. 

On the other hand, conversational query rewriting (CQR) is a generative task that rewrites the conversational input $X_t$ into a standalone question $q_t^{'}$~\cite{yu2020fewshot, voskarides2020quretec, lin2021multistage, kumar2020making, anantha2021open, wu2021conqrr}. Then, existing retrieval systems such as BM25 take the standalone question $q_t^{'}$ to find $P_t$ at inference time, i.e. $f(q_t^{'}, \mathcal{C}) \rightarrow P_t$. As a result, the CQR approaches do not require to re-train additional retriever in a conversational manner. However, the approach is limited in triggering information loss and long latency while rewriting the conversation into the standalone question.

To overcome the limitations, \citet{yu2021few, lin2021contextualized} attempt to train dense retrievers to directly represent the multi-round questions into a single dense vector. They usually focused on few-shot adaptation or weak supervision utilizing other accessible resources including the standalone questions for hard negative mining.

\section{Retrieval Shortcut}
\label{sec:shortcut}
\begin{figure*}[t] 
\centering
\includegraphics[width=1.0\textwidth]{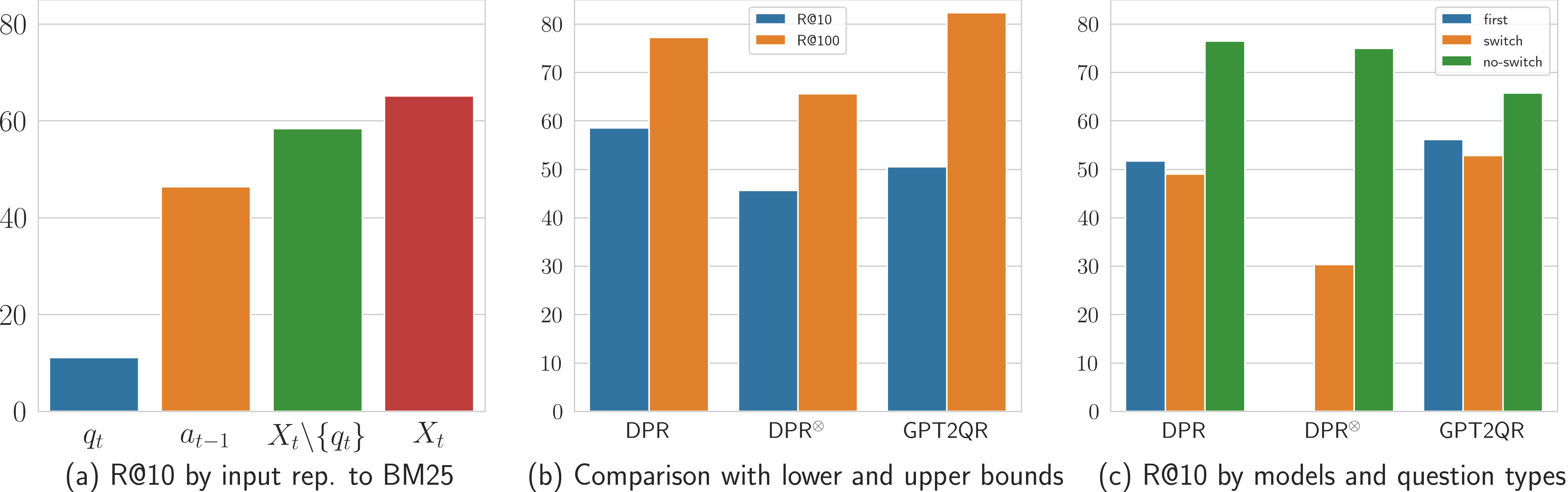}
\caption{Analysis on QReCC~\cite{anantha2021open} for identifying the shortcut. We denote $X_t$ as a conversational input including the current question $q_t$ while $X_t{\setminus}\{q_t\}$ does not contain the $q_t$. (a) Lexical assessment using BM25 to quantify the shortcut in the dataset. BM25 shows small performance drop in R@10 even without considering the current question. (i.e. $X_t{\setminus}\{q_t\}$ as an input.) (b) Comparison of original DPR~\cite{karpukhin2020dense} with its lower bound in terms of shortcut dependency, DPR$^{\otimes}$ taking only $X_t{\setminus}\{q_t\}$ and upper bound, GPT2QR generating and using standalone question $q^{'}_t$ to retrieve. DPR$^{\otimes}$ shows comparable performance without using the current question. (c) Breakdown results of each model by three question types in R@10. Most of the performance gain comes from \textit{no-switch} questions on both original and shortcut-dependent DPRs.}
\label{fig:shortcuts}
\vspace{-3mm}
\end{figure*}

First, we demonstrate the presence of the shortcut in CS datasets. Formally, we define the shortcut as where gold passage $p_t^+$ can be predicted in top-k predictions even without the current question $q_t$. Then, we show how heavily dense retriever relies on the shortcut and how its overall performance is overestimated.

\subsection{Lexical Analysis}

We investigate whether there are spurious \textit{lexical cues} to predict relevant gold passages in CS. Specifically, we input $X_t{\setminus}\{q_t\} = \{q_1, a_1, ..., a_{t-1}\}$ to the BM25 to measure the shortcut. Figure~\ref{fig:shortcuts} (a) shows the result. Surprisingly, we observe the BM25 taking $X_t{\setminus}\{q_t\}$ achieves 58.4 for R@10 on QReCC~\cite{anantha2021open} even without the current question $q_t$. It retains about 90\% of its original performance from BM25 ($X_t$ as an input), indicating $X_t{\setminus}\{q_t\}$ contains enough lexical cues to predict $p_t^+$. However, a model taking only current question $q_t$ does not predict the gold passage well since it does not contain enough lexical cues. Instead, the previous answer $a_{t-1}$ is more responsible for the performance, achieving 46.4 of R@10.

\subsection{Lower and Upper bounds Analysis}

To examine how dense retriever trained on the dataset behave, we contrast a dense retriever with its lower and upper bound models in terms of dependency on the retrieval shortcut. For this, we train two Dense Passage Retriever (DPR) models with in-batch negatives~\cite{karpukhin2020dense} by feeding $X_t$ and $X_t{\setminus}\{q_t\}$ as input query to each model.\footnote{Please see \textsection~\ref{sec:training} for the training details.} We denote the latter one as DPR$^{\otimes}$, and it represents the lower bound model that does not consider the current question $q_t$ at all. Surprisingly, we find the DPR$^{\otimes}$ performs 78\% of R@10 and 85\% of R@100 compared to DPR as shown in Figure~\ref{fig:shortcuts} (b). Thus, we presume the original DPR model is also likely to depend on the shortcut. Next, we introduce the upper bound model, GPT2QR~\cite{anantha2021open}. It is less likely to be exposed to the shortcut since it first generates standalone question $q_t^{'}$, and then its BM25 retriever only takes the decontextualized $q_t^{'}$ as input. We also find that the DPR$^{\otimes}$ is comparable with GPT2QR in R@10 despite the heavy shortcut dependency. It reminds us the overall score is not enough to identify robust retrieval methods.

\subsection{Breakdown by Question types}

To probe when and how models take the shortcut, we break down the evaluation results by question types as in \citet{wu2021conqrr}. Specifically, we define three question types, \textit{first}, \textit{no-switch}, and \textit{switch}. The \textit{first} question is literally first question of conversation without any history. The \textit{no-switch} and \textit{switch} questions can be distinguished by whether $p_{t-1}^+$ contains similar or same topics as $p_t^+$, where the $p_t^+$ is a gold passage at turn $t$ and $t > 1$.\footnote{More details for each question type are in Appendix~\ref{appendix.qtype}.}

Figure~\ref{fig:shortcuts} (c) shows the breakdown result of R@10. The DPR$^{\otimes}$ achieves competitive performance with the DPR in \textit{no-switch} questions, which can benefit from previous conversational history. However, the performances in other two types, \textit{first} and \textit{switch}, drop significantly. Similarly, when we compare DPR with the GPT2QR, we find the performance at \textit{no-switch} turn largely contributes to the gain while degraded in \textit{first} and \textit{switch} types.
As a result, its overreliance on the shortcut prevents the model from generalizing toward real scenarios where a large proportion of topic-switching questions could appear~\cite{adlakha2022topiocqa}. Thus, we claim that the proper ways to take the shortcut could improve the overall score with performance gains at the \textit{first} and \textit{switch} turns while keeping them at the \textit{no-switch}.

\section{Experiments}

We hypothesize random in-batch negatives promote the shortcut dependency of the vanilla DPR model because of their easy-to-distinguish nature. Thus, we examine hard negative mining as one of the solutions to push the retriever to exploit the shortcut properly. We mainly evaluate it on two CS benchmarks, QReCC and OR-QuAC~\cite{anantha2021open, qu2020open}.\footnote{More details of dataset are in Appendix~\ref{appendix.dataset}.}

\subsection{Training Dense Retriever}
\label{sec:training}

DPR consists of two encoders, $E_Q$ and $E_P$, for encoding conversational input and passages, respectively. Each encoders takes the $X_t$ and $p$, a passage in the $\mathcal{C}$, to represent $d$ dimensional vector. Then, we can compute the similarity between the representations via dot product.
\begin{equation*}
    sim(X_t, p) = E_Q(X_t)^T E_P(p)
\end{equation*}

Given the input $X_t$, the encoders are trained in a contrastive manner with the negative passages $P_t^- = \{p_{t1}^-, p_{t2}^-, ..., p_{t|P^-|}^-\}$ and its corresponding positive passage $p_t^+$. 
\begin{equation*}
    L = -log\frac{e^{sim(X_t, p_t^+)}}{e^{sim(X_t, p_t^+)} + \sum_j e^{sim(X_t, p_{tj}^-)}}
\end{equation*}

Basically, we adopt in-batch negatives to obtain the $P_t^-$~\cite{karpukhin2020dense}. For each query representation, it computes the similarity score with other ($B - 1$) number of passage representations except for its gold relevant passage in the same batch, where the $B$ is batch size.

\subsection{Hard Negative Mining}

The in-batch negative is one of the intuitive options to construct the negative examples while reducing memory consumption. However, it is often easy to distinguish from the $p_t^+$ and consequently encourages shortcut dependency. To reduce the dependency, we include a hard negative passage $p^{-}_{t*}$ in the $P_t^{-}$.
We first construct $k$ number of negative passages $N_t^{-}$ for each training instance. Then, we randomly sample a passage from the $N_t^{-}$ to include it in $P_t^{-}$ as the $p^{-}_{t*}$. We denote off-the-shelf retriever to obtain top-$k$ passages in $\mathcal{C}$ from given input query $x$ as $\mathcal{F}(x, \mathcal{C}, k)$. 
Specifically, we compare three strategies for hard negative mining:

\paragraph{BM25 Negs} De-facto strategy is BM25-based negative mining following \citet{karpukhin2020dense}. We mine the $N_t^{-}$ using whole conversational input $X_t$, i.e., $N_t^{-} \leftarrow \text{BM25}(X_t, \mathcal{C}, k)$.

\paragraph{CQR Negs} If gold standalone question $q_t^{'}$ is available for each $X_t$, we can leverage it to find the negative passages with off-the-shelf retriever as in ~\citet{yu2020fewshot, lin2021contextualized}, i.e., $N_t^{-} \leftarrow \mathcal{F}(q_t^{'}, \mathcal{C}, k)$. For this, we employ another DPR pre-trained on Natural Questions (NQ)~\cite{kwiatkowski2019natural} as the $\mathcal{F}$.

\paragraph{Model Negs} Lastly, we explore model-based hard negative mining proposed by \citet{xiong2020approximate}. First, we train vanilla DPR model on the target dataset using only in-batch negative as in \textsection ~\ref{sec:shortcut}. Then, we employ the model as $\mathcal{F}$ to select negative passages, i.e., $N_t^{-} \leftarrow \mathcal{F}(X_t, \mathcal{C}, k)$.

\subsection{Implementation Details}

DPR pre-trained on NQ dataset~\cite{kwiatkowski2019natural} of \citet{karpukhin2020dense} is used for the initial checkpoint of our dense retrievers. It consists of two BERT encoders and 220M of learnable parameters~\cite{devlin2019bert}. We set maximum sequence length to 128 and 384 for $X_t$ and $p$, respectively. All history is concatenated with a \texttt{[SEP]} token in between. We retrain the first question and truncate tokens from the left side up to the maximum length of 128 for $X_t$.

We train the models for 10 epochs with 3e-5 of learning rate (lr). For optimization, AdamW is used with 0.1 warming up ratio for linear lr decay scheduling~\cite{kingma2017adam}. We build top 100 passages for the hard negatives $N_t^{-}$, i.e., $k=100$. Batch size is set to 128 for OR-QuAC and 256 for QReCC. We choose the best performing model based on dev set. We use Pyserini~\cite{lin2021pyserini} to implement BM25 and IndexFlatIP index of FAISS~\cite{johnson2019billion} to perform dense retrieval.\footnote{All our experiments is based on NSML platform~\cite{sung2017nsml,kim2018nsml} and Transformers library~\cite{wolf2020transformers} using \{4, 8\} 32GB V100 GPUs.}

\begin{table*}[t!]
    \centering
    \small
    \setlength{\tabcolsep}{0.4em}
    \begin{threeparttable}
    \begin{tabular*}{0.95\textwidth}{lcccccccccccc}
        \toprule
        & \multicolumn{3}{c}{All} & \multicolumn{3}{c}{first} & \multicolumn{3}{c}{switch} & \multicolumn{3}{c}{no-switch} \\
        \cmidrule(lr){2-4} \cmidrule(lr){5-7} \cmidrule(lr){8-10} \cmidrule(lr){11-13}
        Model & MRR & R@10 & R@100 & MRR & R@10 & R@100 & MRR & R@10 & R@100 & MRR & R@10 & R@100 \\
        \midrule
        BM25 & 0.47 & 65.1 & 82.8 & 0.32 & 56.1 & \textbf{99.1} & 0.18 & 36.3 & 70.5 & \textbf{0.78} & \textbf{90.8} & \textbf{97.9} \\
        BM25$^{\otimes}$ & 0.43 & 58.4 & 63.9 & - & - & - & 0.16 & 32.8 & 65.3 & 0.76 & 90.3 & 96.5 \\
        DPR$^{\otimes}$ & 0.28 & 46.5 & 65.9 & - & - & - & 0.14 & 30.2 & 56.7 & 0.54 & 75.5 & 88.4 \\
        GPT2QR & 0.32 & 50.5 & 82.3 & 0.32 & 56.1 & \textbf{99.1} & 0.30 & 52.8 & 88.0 & 0.46 & 65.7 & 88.9 \\
        CONQRR & 0.42 & 65.1 & 84.7 & - & - & - & - & - & - & - & - & - \\
        \midrule
        DPR & 0.39 & 59.1 & 77.6 & 0.36 & 55.7 & 77.6 & 0.29 & 50.3 & 71.5 & 0.60 & 80.8 & 90.8  \\
        w. CQR Negs & 0.50 & 71.6 & 86.0 & 0.42 & 64.0 & 82.2 & 0.34 & 57.9 & 80.8 & 0.70 & 86.7 & 94.2 \\
        w. BM25 Negs & 0.51 & 73.5 & 86.3 & 0.46 & 65.5 & 85.7 & 0.38 & 61.8 & 82.3 & 0.68 & 85.5 & 92.9 \\
        w. Model Negs & \textbf{0.53} & \textbf{76.1} & \textbf{88.3} & \textbf{0.48} & \textbf{70.9} & 87.1 & \textbf{0.40} & \textbf{63.0} & \textbf{84.1} & 0.72 & 88.1 & 94.1  \\
        \bottomrule
    \end{tabular*}
    \end{threeparttable}
    \caption{Experimental results on QReCC test set (All) and its sub-splits by three question types discussed in \textsection~\ref{sec:shortcut}. The $^{\otimes}$ indicates the model takes only $X_t{\setminus}\{q_t\}$ as input.}
    \label{table:main_qrecc}
\vspace{-3mm}
\end{table*}

\subsection{Baselines}

In QReCC, we include BM25 and BM25$^\otimes$ take $X_t$ and $X_t{\setminus}\{q_t\}$ as input query, respectively. For CQR baselines less dependent on the shortcut, we include GPT2QR and CONQRR~\cite{anantha2021open, wu2021conqrr}. They use standalone question instead of directly encoding a conversation for the input of off-the-shelf retriever such as BM25 or T5-DE~\cite{ni-etal-2022-sentence} finetuned on ODQA dataset. \citet{anantha2021open} propose GPT2QR as baseline model which is GPT-2~\cite{radford2019language} based CQR model. We only perform BM25 inference based on released model predictions by authors instead of re-training it. CONQRR is based on T5~\cite{raffel2020exploring} for the CQR~\cite{wu2021conqrr}. Especially, \citet{wu2021conqrr} train the CONQRR using reinforcement learning against retrieval metrics (MRR, Recall) as reward signals. We also include DPR and DPR$^\otimes$ without hard negative mining to represent shortcut-dependent model.

In OR-QuAC, we compare our models with previously proposed dense retrieval approaches in conversational search, CQE~\cite{lin2021contextualized} and ConvDR~\cite{yu2020fewshot}. Both of them utilize the standalone question $q_t^{'}$ to mine hard negatives and knowledge distillation from off-the-shelf retrievers trained on ODQA, regarding it as a teacher model. Although they were not tested on QReCC, we can indirectly compare them with others using DPR with CQR Negs instead.

\subsection{Results}

We report scores among Mean Reciprocal Rank (MRR) and Recall (R@K, K $\in \{5, 10, 100\}$).\footnote{We use \href{https://github.com/scai-conf/SCAI-QReCC-21/}{updated evaluation script}, which does not consider when there is no ground truth, by \citet{vakulenko2022scai}.} Table~\ref{table:main_qrecc} shows the retrieval performances of baseline models and hard negative mining methods on QReCC, and our findings are summarized:

\textbf{Overall performances are not enough to distinguish robust methods in CS.} We find lexical baselines, BM25 and BM25$^{\otimes}$, outperform CQR-based models, GPT2QR and CONQRR~\cite{anantha2021open, wu2021conqrr} and vanilla DPR in MRR of overall retrieval performances (All). However, as we discussed in \textsection ~\ref{sec:shortcut}, the most performances are from \textit{no-switch} questions which can benefit from the shortcut.

\begin{figure}
     \centering
     \includegraphics[width=\linewidth]{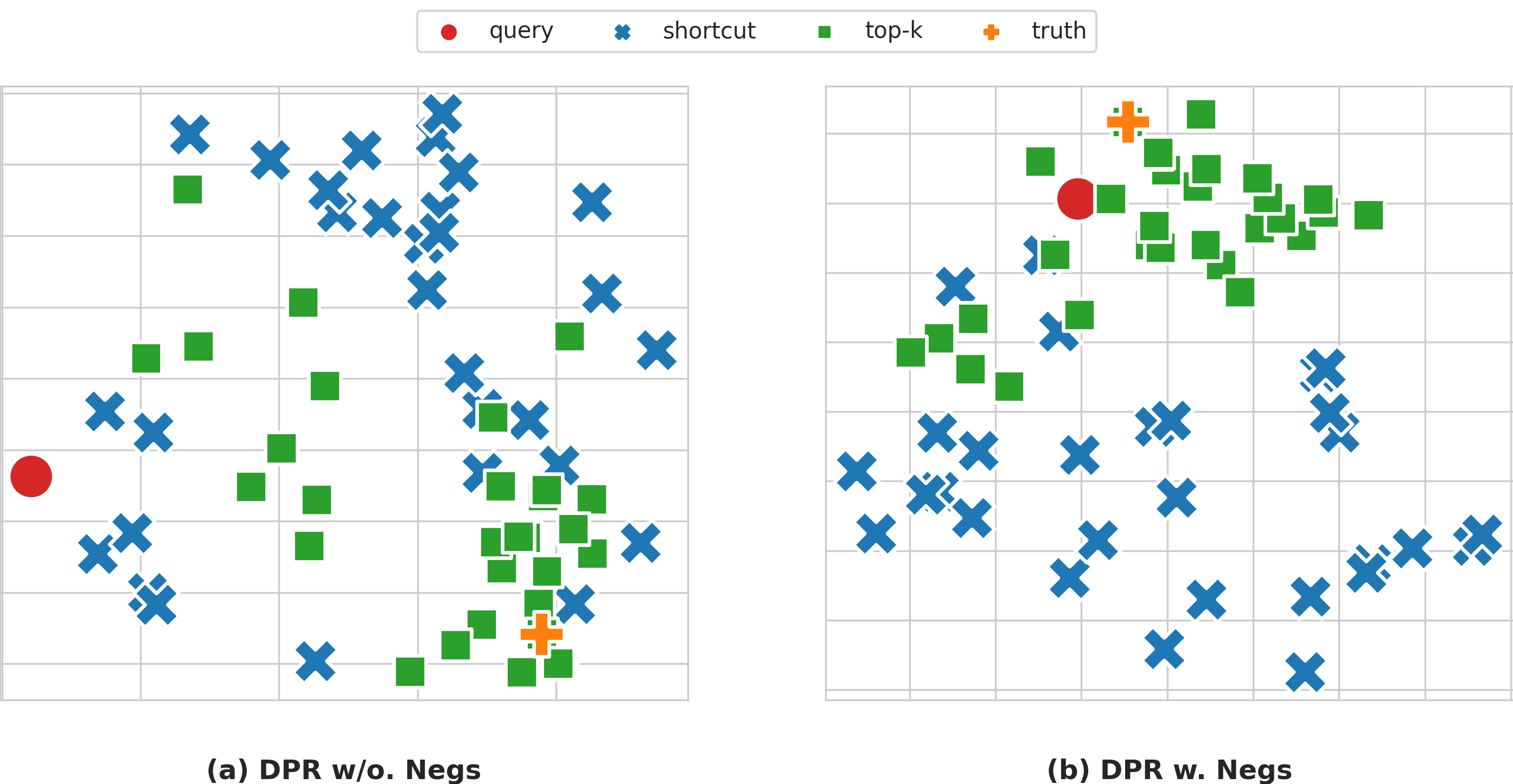}
        \caption{T-SNE visualization of query and passage embeddings based on two DPR models with and without hard negative training. The shortcut (blue multiply) passages are obtained by BM25$^{\otimes}$. The example is from 5th turn of conversation 1935 in QReCC test set, which is one of \textit{switch} questions. Please see Appendix~\ref{appendix.qualitative} for the corresponding qualitative example.}
        \label{fig:tsne}
        \vspace{-3mm}
\end{figure}

\textbf{Hard negatives could mitigate shortcut dependency of dense retrievers.} We observe the vanilla DPR underperforms the GPT2QR in \textit{first} and \textit{switch} questions. Also, there is a relatively smaller gap between DPR$^{\otimes}$ and DPR in \textit{no-switch} type of questions. Compared to the vanilla DPR, all three negatives effectively improve the overall performance. Especially, the history-independent types, \textit{first} and \textit{switch}, are improved at most 12.7-15.2 in R@10 indicating relaxed shortcut dependency of the model. Figure~\ref{fig:tsne} shows T-SNE visualizations~\cite{van2008visualizing} to compare DPR models with and without negatives in embedding space. Different from the vanilla DPR that fails to identify a gold passage from other irrelevant passages, DPR trained with negatives more clearly discriminates it from the distractors.

\textbf{Among the negative mining methods, the model-based hard negative consistently outperforms others}. We observe consistent results in other CS dataset, OR-QuAC~\cite{qu2020open} compared to previous works (Please see Appendix~\ref{appendix.orquac}). Moreover, our model achieves a new state-of-the-art with improvements of 11.0\% point R@10.

\section{Conclusion}
In this work, we show the presence of the shortcut in conversational search, which causes dense retriever often heavily relies on it when trained on in-batch negatives. We find that shortcut dependency hurts the generalization ability of dense retrievers. To save the model from relying on the shortcut, we study various hard negative mining strategies. The retriever trained with hard negatives appropriately takes beneficial information of the shortcut only when needed and achieves the state-of-the-art performance on multiple CS benchmarks.

\section*{Limitations}
Even if we explain the existence of shortcut in conversational search, we could not suggest specific solutions to the shortcut dependency of dense retrievers. In the preliminary study, we tried other methods, e.g., history masking to promote the model attending more to the current question, but we found those methods are not effective as the hard negative mining in terms of shortcut dependency.
However, we believe our work is an important step toward more robust conversational search.

Another limitation is the implementation cost to perform the model-based hard negative mining, i.e., indexing and inference of dense retriever over huge passages collection. Please see Appendix~\ref{appendix.computational_cost} for the details. Especially, the cost is increased notoriously according to the number of passage collections. We expect a more efficient method to balance shortcut dependency in future works.

\section*{Acknowledgements}

The authors would like to thank to Kyunghyun Cho, Jinhyuk Lee, Minjoon Seo, Sang-Woo Lee, Hwaran Lee and other members of NAVER AI for their constructive comments.

\bibliography{custom}

\begin{thebibliography}{30}
\expandafter\ifx\csname natexlab\endcsname\relax\def\natexlab#1{#1}\fi

\bibitem[{Adlakha et~al.(2022)Adlakha, Dhuliawala, Suleman, de~Vries, and
  Reddy}]{adlakha2022topiocqa}
Vaibhav Adlakha, Shehzaad Dhuliawala, Kaheer Suleman, Harm de~Vries, and Siva
  Reddy. 2022.
\newblock Topiocqa: Open-domain conversational question answering with topic
  switching.
\newblock \emph{Transactions of the Association for Computational Linguistics},
  10:468--483.

\bibitem[{Anantha et~al.(2021)Anantha, Vakulenko, Tu, Longpre, Pulman, and
  Chappidi}]{anantha2021open}
Raviteja Anantha, Svitlana Vakulenko, Zhucheng Tu, Shayne Longpre, Stephen
  Pulman, and Srinivas Chappidi. 2021.
\newblock Open-domain question answering goes conversational via question
  rewriting.
\newblock In \emph{Proceedings of the 2021 Conference of the North American
  Chapter of the Association for Computational Linguistics: Human Language
  Technologies (NAACL)}, pages 520--534.

\bibitem[{Chen et~al.(2017)Chen, Fisch, Weston, and Bordes}]{chen2017reading}
Danqi Chen, Adam Fisch, Jason Weston, and Antoine Bordes. 2017.
\newblock Reading wikipedia to answer open-domain questions.
\newblock In \emph{Proceedings of the 55th Annual Meeting of the Association
  for Computational Linguistics (Volume 1: Long Papers) (ACL)}, pages
  1870--1879.

\bibitem[{Choi et~al.(2018)Choi, He, Iyyer, Yatskar, Yih, Choi, Liang, and
  Zettlemoyer}]{choi2018quac}
Eunsol Choi, He~He, Mohit Iyyer, Mark Yatskar, Wen-tau Yih, Yejin Choi, Percy
  Liang, and Luke Zettlemoyer. 2018.
\newblock Quac: Question answering in context.
\newblock In \emph{Proceedings of the 2018 Conference on Empirical Methods in
  Natural Language Processing (EMNLP)}, pages 2174--2184.

\bibitem[{Dalton et~al.(2019)Dalton, Xiong, and Callan}]{dalton2019cast}
Jeffrey Dalton, Chenyan Xiong, and Jamie Callan. 2019.
\newblock Cast 2019: The conversational assistance track overview.
\newblock In \emph{Proceedings of the Twenty-Eighth Text REtrieval Conference,
  TREC}, pages 13--15.

\bibitem[{Devlin et~al.(2019)Devlin, Chang, Lee, and
  Toutanova}]{devlin2019bert}
Jacob Devlin, Ming-Wei Chang, Kenton Lee, and Kristina Toutanova. 2019.
\newblock Bert: Pre-training of deep bidirectional transformers for language
  understanding.
\newblock In \emph{In North American Chapter of the Association for
  Computational Linguistics (NAACL).}

\bibitem[{Elgohary et~al.(2019)Elgohary, Peskov, and
  Boyd-Graber}]{elgohary2019can}
Ahmed Elgohary, Denis Peskov, and Jordan Boyd-Graber. 2019.
\newblock Can you unpack that? learning to rewrite questions-in-context.
\newblock In \emph{Proceedings of the 2019 Conference on Empirical Methods in
  Natural Language Processing and the 9th International Joint Conference on
  Natural Language Processing (EMNLP-IJCNLP)}, pages 5918--5924.

\bibitem[{Johnson et~al.(2019)Johnson, Douze, and
  J{\'e}gou}]{johnson2019billion}
Jeff Johnson, Matthijs Douze, and Herv{\'e} J{\'e}gou. 2019.
\newblock Billion-scale similarity search with gpus.
\newblock \emph{IEEE Transactions on Big Data}, 7(3):535--547.

\bibitem[{Karpukhin et~al.(2020)Karpukhin, Oguz, Min, Lewis, Wu, Edunov, Chen,
  and Yih}]{karpukhin2020dense}
Vladimir Karpukhin, Barlas Oguz, Sewon Min, Patrick Lewis, Ledell Wu, Sergey
  Edunov, Danqi Chen, and Wen-tau Yih. 2020.
\newblock Dense passage retrieval for open-domain question answering.
\newblock In \emph{Proceedings of the 2020 Conference on Empirical Methods in
  Natural Language Processing (EMNLP)}, pages 6769--6781.

\bibitem[{Kim et~al.(2018)Kim, Kim, Seo, Kim, Park, Park, Jo, Kim, Yang, Kim
  et~al.}]{kim2018nsml}
Hanjoo Kim, Minkyu Kim, Dongjoo Seo, Jinwoong Kim, Heungseok Park, Soeun Park,
  Hyunwoo Jo, KyungHyun Kim, Youngil Yang, Youngkwan Kim, et~al. 2018.
\newblock Nsml: Meet the mlaas platform with a real-world case study.
\newblock \emph{arXiv preprint arXiv:1810.09957}.

\bibitem[{Kingma and Ba(2017)}]{kingma2017adam}
Diederik~P. Kingma and Jimmy Ba. 2017.
\newblock \href {http://arxiv.org/abs/1412.6980} {Adam: A method for stochastic
  optimization}.

\bibitem[{Kumar and Callan(2020)}]{kumar2020making}
Vaibhav Kumar and Jamie Callan. 2020.
\newblock Making information seeking easier: An improved pipeline for
  conversational search.
\newblock In \emph{Findings of the Association for Computational Linguistics:
  EMNLP 2020}, pages 3971--3980.

\bibitem[{Kwiatkowski et~al.(2019)Kwiatkowski, Palomaki, Redfield, Collins,
  Parikh, Alberti, Epstein, Polosukhin, Devlin, Lee
  et~al.}]{kwiatkowski2019natural}
Tom Kwiatkowski, Jennimaria Palomaki, Olivia Redfield, Michael Collins, Ankur
  Parikh, Chris Alberti, Danielle Epstein, Illia Polosukhin, Jacob Devlin,
  Kenton Lee, et~al. 2019.
\newblock Natural questions: A benchmark for question answering research.
\newblock \emph{Transactions of the Association for Computational Linguistics
  (TACL)}, 7:452--466.

\bibitem[{Lin et~al.(2021{\natexlab{a}})Lin, Ma, Lin, Yang, Pradeep, and
  Nogueira}]{lin2021pyserini}
Jimmy Lin, Xueguang Ma, Sheng-Chieh Lin, Jheng-Hong Yang, Ronak Pradeep, and
  Rodrigo Nogueira. 2021{\natexlab{a}}.
\newblock Pyserini: An easy-to-use python toolkit to support replicable ir
  research with sparse and dense representations.
\newblock \emph{arXiv preprint arXiv:2102.10073}.

\bibitem[{Lin et~al.(2021{\natexlab{b}})Lin, Yang, and
  Lin}]{lin2021contextualized}
Sheng-Chieh Lin, Jheng-Hong Yang, and Jimmy Lin. 2021{\natexlab{b}}.
\newblock Contextualized query embeddings for conversational search.
\newblock In \emph{Proceedings of the 2021 Conference on Empirical Methods in
  Natural Language Processing (EMNLP)}, pages 1004--1015.

\bibitem[{Lin et~al.(2021{\natexlab{c}})Lin, Yang, Nogueira, Tsai, Wang, and
  Lin}]{lin2021multistage}
Sheng-Chieh Lin, Jheng-Hong Yang, Rodrigo Nogueira, Ming-Feng Tsai, Chuan-Ju
  Wang, and Jimmy Lin. 2021{\natexlab{c}}.
\newblock Multi-stage conversational passage retrieval: An approach to fusing
  term importance estimation and neural query rewriting.
\newblock \emph{ACM Transactions on Information Systems (TOIS)}, 39(4):1--29.

\bibitem[{Ni et~al.(2022)Ni, Hernandez~Abrego, Constant, Ma, Hall, Cer, and
  Yang}]{ni-etal-2022-sentence}
Jianmo Ni, Gustavo Hernandez~Abrego, Noah Constant, Ji~Ma, Keith Hall, Daniel
  Cer, and Yinfei Yang. 2022.
\newblock Sentence-t5: Scalable sentence encoders from pre-trained text-to-text
  models.
\newblock In \emph{Findings of the Association for Computational Linguistics:
  ACL 2022}, Dublin, Ireland. Association for Computational Linguistics.

\bibitem[{Qu et~al.(2020)Qu, Yang, Chen, Qiu, Croft, and Iyyer}]{qu2020open}
Chen Qu, Liu Yang, Cen Chen, Minghui Qiu, W~Bruce Croft, and Mohit Iyyer. 2020.
\newblock Open-retrieval conversational question answering.
\newblock In \emph{Proceedings of the 43rd International ACM SIGIR conference
  on research and development in Information Retrieval (SIGIR)}, pages
  539--548.

\bibitem[{Radford et~al.(2019)Radford, Wu, Child, Luan, Amodei, Sutskever
  et~al.}]{radford2019language}
Alec Radford, Jeffrey Wu, Rewon Child, David Luan, Dario Amodei, Ilya
  Sutskever, et~al. 2019.
\newblock Language models are unsupervised multitask learners.
\newblock \emph{OpenAI blog}, 1(8):9.

\bibitem[{Raffel et~al.(2020)Raffel, Shazeer, Roberts, Lee, Narang, Matena,
  Zhou, Li, Liu et~al.}]{raffel2020exploring}
Colin Raffel, Noam Shazeer, Adam Roberts, Katherine Lee, Sharan Narang, Michael
  Matena, Yanqi Zhou, Wei Li, Peter~J Liu, et~al. 2020.
\newblock Exploring the limits of transfer learning with a unified text-to-text
  transformer.
\newblock \emph{J. Mach. Learn. Res. (JMLR)}, 21(140):1--67.

\bibitem[{Sung et~al.(2017)Sung, Kim, Jo, Yang, Kim, Lausen, Kim, Lee, Kwak, Ha
  et~al.}]{sung2017nsml}
Nako Sung, Minkyu Kim, Hyunwoo Jo, Youngil Yang, Jingwoong Kim, Leonard Lausen,
  Youngkwan Kim, Gayoung Lee, Donghyun Kwak, Jung-Woo Ha, et~al. 2017.
\newblock Nsml: A machine learning platform that enables you to focus on your
  models.
\newblock \emph{arXiv preprint arXiv:1712.05902}.

\bibitem[{Vakulenko et~al.(2022)Vakulenko, Kiesel, and
  Fr{\"o}be}]{vakulenko2022scai}
Svitlana Vakulenko, Johannes Kiesel, and Maik Fr{\"o}be. 2022.
\newblock Scai-qrecc shared task on conversational question answering.
\newblock \emph{arXiv preprint arXiv:2201.11094}.

\bibitem[{Van~der Maaten and Hinton(2008)}]{van2008visualizing}
Laurens Van~der Maaten and Geoffrey Hinton. 2008.
\newblock Visualizing data using t-sne.
\newblock \emph{Journal of machine learning research (JMLR)}, 9(11).

\bibitem[{Voorhees and Tice(2000)}]{voorhees-tice-2000-trec}
Ellen~M. Voorhees and Dawn~M. Tice. 2000.
\newblock \href {http://www.lrec-conf.org/proceedings/lrec2000/pdf/26.pdf} {The
  {TREC}-8 question answering track}.
\newblock In \emph{Proceedings of the Second International Conference on
  Language Resources and Evaluation ({LREC}{'}00)}, Athens, Greece. European
  Language Resources Association (ELRA).

\bibitem[{Voskarides et~al.(2020)Voskarides, Li, Ren, Kanoulas, and
  de~Rijke}]{voskarides2020quretec}
Nikos Voskarides, Dan Li, Pengjie Ren, Evangelos Kanoulas, and Maarten
  de~Rijke. 2020.
\newblock Query resolution for conversational search with limited supervision.
\newblock In \emph{Proceedings of the 43rd International ACM SIGIR conference
  on research and development in Information Retrieval (SIGIR)}, pages
  921--930.

\bibitem[{Wolf et~al.(2020)Wolf, Debut, Sanh, Chaumond, Delangue, Moi, Cistac,
  Rault, Louf, Funtowicz et~al.}]{wolf2020transformers}
Thomas Wolf, Lysandre Debut, Victor Sanh, Julien Chaumond, Clement Delangue,
  Anthony Moi, Pierric Cistac, Tim Rault, Remi Louf, Morgan Funtowicz, et~al.
  2020.
\newblock Transformers: State-of-the-art natural language processing.
\newblock In \emph{Proceedings of the 2020 Conference on Empirical Methods in
  Natural Language Processing: System Demonstrations (EMNLP demo)}, pages
  38--45.

\bibitem[{Wu et~al.(2021)Wu, Luan, Rashkin, Reitter, and Tomar}]{wu2021conqrr}
Zeqiu Wu, Yi~Luan, Hannah Rashkin, David Reitter, and Gaurav~Singh Tomar. 2021.
\newblock Conqrr: Conversational query rewriting for retrieval with
  reinforcement learning.
\newblock \emph{arXiv preprint arXiv:2112.08558}.

\bibitem[{Xiong et~al.(2020)Xiong, Xiong, Li, Tang, Liu, Bennett, Ahmed, and
  Overwijk}]{xiong2020approximate}
Lee Xiong, Chenyan Xiong, Ye~Li, Kwok-Fung Tang, Jialin Liu, Paul~N Bennett,
  Junaid Ahmed, and Arnold Overwijk. 2020.
\newblock Approximate nearest neighbor negative contrastive learning for dense
  text retrieval.
\newblock In \emph{International Conference on Learning Representations
  (ICLR)}.

\bibitem[{Yu et~al.(2020)Yu, Liu, Yang, Xiong, Bennett, Gao, and
  Liu}]{yu2020fewshot}
Shi Yu, Jiahua Liu, Jingqin Yang, Chenyan Xiong, Paul Bennett, Jianfeng Gao,
  and Zhiyuan Liu. 2020.
\newblock Few-shot generative conversational query rewriting.
\newblock In \emph{Proceedings of the 43rd International ACM SIGIR conference
  on research and development in Information Retrieval (SIGIR)}, pages
  1933--1936.

\bibitem[{Yu et~al.(2021)Yu, Liu, Xiong, Feng, and Liu}]{yu2021few}
Shi Yu, Zhenghao Liu, Chenyan Xiong, Tao Feng, and Zhiyuan Liu. 2021.
\newblock Few-shot conversational dense retrieval.
\newblock In \emph{Proceedings of the 44th International ACM SIGIR Conference
  on Research and Development in Information Retrieval (SIGIR)}, pages
  829--838.

\end{thebibliography}
\bibliographystyle{acl_natbib}

\clearpage
\appendix
\label{sec:appendix}

\section{Details of Question Types}
\label{appendix.qtype}
We classify the \textit{no-switch} and \textit{switch} questions using dot product score between BM25 vectors of $p_{t-1}^+$ and $p_{t}^+$ as threshold in QReCC dataset. This is similar with division of \textit{topic-concentrated} and \textit{topic-shifted} questions in \citet{wu2021conqrr} while we take them only when $t>1$ to distinguish them from \textit{first} questions. The number of subsets is 267, 279, and 573 for the \textit{first}, \textit{no-switch}, and \textit{switch} respectively. Please note that the sum of each subset is not equal to the number of \textit{all} (8209) since we take the question types from only NQ and TREC subdomains in the QReCC dataset as in \citet{wu2021conqrr}. 

\section{Details of Dataset}
\label{appendix.dataset}

\begin{table}[h]
    \small
    \centering
    \begin{threeparttable}
    \begin{tabular*}{\columnwidth}{llrrrc}
        \toprule
        Dataset & & Train & Dev & Test & $\mathcal{C}$ \\
        \midrule
        \multirow{2}{*}{\textbf{OR-QuAC}} & \# C & 4,383 & 490 & 771 & \multirow{2}{*}{11M}\\
        & \# Q & 31,526 & 3,430 & 5,571 & \\
        \midrule
        \multirow{2}{*}{\textbf{QReCC}} & \# C & 8,823 & 2,000 & 2,775 & \multirow{2}{*}{54M}\\
        & \# Q & 51,928 & 11,573 & 16,451 & \\
        \bottomrule
    \end{tabular*}
    \end{threeparttable}
    \caption{Dataset statistics used in our experiments. The \# C and \# Q indicate the number of conversations and questions, respectively.}
    \label{table:data_stats}
\end{table}

We mainly conduct experiments on recent CS benchmarks, OR-QuAC and QReCC~\cite{qu2020open, anantha2021open}. We briefly describe the procedures of data construction and features of each dataset. Table~\ref{table:data_stats} shows dataset statistics we used.

 \paragraph{OR-QuAC} \citet{qu2020open} extend one of the popular CQA datasets, QuAC~\cite{choi2018quac} to the open-domain setting by aligning relevant passages with the questions in QuAC.~\footnote{\href{https://github.com/prdwb/orconvqa-release}{github.com/prdwb/orconvqa-release}} Moreover, they facilitate CQR as a subtask by reusing examples in CANARD~\cite{elgohary2019can}. For retrieval, they construct passage collections from Wikipedia. However, the dataset has limitations in that all questions in the same conversation share the same gold passage. In other words, most of the questions in OR-QuAC are \textit{no-switch} type. Thus, it is vulnerable to the shortcut. Even though it is far from the real world scenario, we include OR-QuAC to compare previous dense retrieval approaches~\cite{lin2021contextualized, yu2021few}. We use smaller collections $\mathcal{C}_{dev}$ (6.9k) provided by the authors for the development.
 
  \paragraph{QReCC} \citet{anantha2021open} construct QReCC dataset based on three existing datasets, QuAC, Natural Questions (NQ), and TREC~\cite{choi2018quac, kwiatkowski2019natural, dalton2019cast}.~\footnote{\href{https://zenodo.org/record/5115890\#.YgCWNfVBxhF}{zenodo.org/record/5115890\#.YgCWNfVBxhF}} To annotate gold passage, they reuse conversational questions in QuAC and CAsT as in \citet{qu2020open}, while collecting new questions for the NQ dataset. Given a question randomly selected from NQ, each crowdworker alone generates not only the following questions but also their corresponding answers. Even though it contains more diverse and realistic questions than the OR-QuAC, most of the questions (78\%) still belong to the QuAC, causing models to exploit the shortcut. We newly select the development set by sampling 2k conversations from the train set, since \citet{anantha2021open} combined them into the train set when the dataset is released. We also choose 7.3k number of corresponding dev passages for the development collections $\mathcal{C}_{dev}$. We only regard the examples that contain ground truth relevant passages. Thus, the actual number of training examples is 24,283.

\section{Experimental Results on OR-QuAC}
\label{appendix.orquac}
\begin{table}[t!]
    \centering
    \begin{threeparttable}
    \begin{tabular*}{0.95\columnwidth}{lcc}
        \toprule
        Model & MRR & R@5 \\
        \midrule
        BM25($q_1, \mathcal{C}$)& 0.216 & 30.6 \\
        BM25($q_t, \mathcal{C}$)& 0.043 & 5.6 \\
        BM25($Q_{t-1}, \mathcal{C}$)& 0.170 & 21.3 \\
        BM25($Q_t, \mathcal{C}$) & 0.198 & 24.9 \\
        \midrule
        ALBERT \cite{qu2020open} & 0.225 & 31.4 \\
        CQE$^{\diamond}$ \cite{lin2021contextualized} & 0.266 & 36.5 \\
        ConvDR \cite{yu2021few} & 0.616 & 75.0 \\
        \midrule
        DPR & 0.525 & 63.9 \\
        w. Model Negs & \textbf{0.633} & \textbf{75.9} \\
        \bottomrule
    \end{tabular*}
    \end{threeparttable}
    \caption{Experimental result on OR-QuAC. Please note that all models take only multi-round questions $Q_t = \{q_1, q_2, ..., q_t\}$ instead of $X_t$ as input following previous works. The $^{\diamond}$ indicates the CQE model performs zero-shot inference and dimensionality reduction~\cite{lin2021contextualized}.}
    \label{table:orquac}
\end{table}

Table~\ref{table:orquac} shows results on OR-QuAC where most of the questions are \textit{no-switch} type. First, we observe another retrieval shortcut on the first question, which is not observed in QReCC. Even if we input only first question to BM25, BM25($q_1, \mathcal{C})$, it achieves competitive results with ALBERT baseline by \citet{qu2020open}. We presume the lexical cues from the first question are caused by pre-proccesing for the questions, rewriting to the standalone questions~\cite{qu2020open}.

Our DPR with model-based hard negatives consistently outperforms the previous dense retrievers~\cite{yu2020fewshot, lin2021contextualized}. Even though it is not fair comparison since their different backbones and setups, we can compare the models in terms of hard negative mining strategies. Both CQE and ConvDR utilize CQR-based negatives requiring gold human rewrite $q_t^{'}$ (CQR Negs). Similar to result in Table~\ref{table:main_qrecc}, our model with model-based negatives (Model Negs) achieves better performances without any usage of query rewriting.

\section{Computational Cost}
\label{appendix.computational_cost}

\begin{table}[t!]
    \centering
    \begin{threeparttable}
    \begin{tabular*}{\columnwidth}{lrrr}
        \toprule
        Data & Training & Indexing & Inference \\
        \midrule
        OR-QuAC & 2h & 8h & 40m \\
        QReCC & 2h & 28h & 11h \\
        \bottomrule
    \end{tabular*}
    \end{threeparttable}
    \caption{Summarized computational cost (run-time) for each training, indexing, and inference of dense retrieval. The target of each function is train set, passages collection, and test or dev set.}
    \label{table:computational}
\end{table}

Overall computational cost is summarized in Table~\ref{table:computational}. Please note that the number of passages collection and test set of QReCC is much larger than the other. Thus, we allocate 8 GPUs for QReCC and 4 GPUs for OR-QuAC to perform training and indexing. We conduct training and inference once for all experiments because of the expensive computational cost.

\clearpage
\onecolumn
\section{Qualitative Examples}
\label{appendix.qualitative}

\begin{table*}[h]
\centering
\resizebox{\textwidth}{!}{
\begin{tabular}{l}
\toprule
\large \textbf{Conversation (id: 1935\_5)}  \\
\midrule
\textbf{$q_\text{1}$} : when was a view to a kill made? \\
\textbf{$a_\text{1}$} : a view to a kill is a 1985 spy film and the fourteenth in the james bond series produced by eon productions \\
\textbf{$q_\text{2}$} : who directed the film? \\
\textbf{$a_\text{2}$} : a view to kill was the third james bond film and was directed by john glen \\
\textbf{$q_\text{3}$} : who played the main character? \\
\textbf{$a_\text{3}$} : roger moore played james bond, mi6 agent 007, in a view to kill \\
\textbf{$q_\text{4}$} : how old was the actor? \\
\textbf{$a_\text{4}$} : roger \colorbox{red!40}{\textbf{moore}} starred in his \colorbox{red!40}{\textbf{final bond film}}, a view to a kill at the age of 58. \\moore was \colorbox{red!40}{\textbf{the oldest actor}} to have played bond \\
\textbf{$q_\text{5}$} : \colorbox{blue!40}{\textbf{in how many films did the actor play the character?}} \\
\midrule
\textbf{DPR without Negatives} \\
\midrule
In the Bond movie universe, \colorbox{red!40}{\textbf{Moore}} is the proud holder of three noteworthy records: \\1. At age 45 in 1973, he was the oldest actor to take over the role.\\ 2. At 57, he was again \colorbox{red!40}{\textbf{the oldest actor play}} the role in 1985’s A View To a Kill, his seventh and \colorbox{red!40}{\textbf{final Bond film}}.\\ 3. With seven appearances as 007, he starred in the most official Bond films. $\cdots$ \\
\midrule
\textbf{DPR with Negatives} \\
\midrule
$\cdots$ British actor Sir \colorbox{blue!40}{\textbf{Roger Moore}} KBE Moore in 1973 Born Roger George Moore ( 1927-10-14 )\\ 14 October 1927 Stockwell, London, England Died 23 May 2017 (2017-05-23) (aged 89) Crans-Montana, \\Switzerland [1] Burial place Monaco Cemetery Alma mater Royal Academy of Dramatic Art Occupation\\ Actor Years active 1945–2013 Known for \colorbox{blue!40}{\textbf{James Bond in seven feature films}} from 1973 to 1985 $\cdots$  \\
\bottomrule
\end{tabular}
}
\caption{An example of top-1 predictions from vanilla DPR (without Negatives) and DPR trained with model-based hard negatives (with Negatives). The vanilla DPR without hard negatives fails to predict a gold passage since it heavily relies on shortcut, i.e., previous answer $a_4$. On the other hand, the DPR successfully predicts a gold passage with comprehending whole conversational context up to $q_5$ when the retriever is trained with hard negatives.}
\label{table:qualitative_example}
\end{table*}

\end{document}